\documentclass[a4paper,fleqn]{cas-sc}

\usepackage[authoryear,longnamesfirst]{natbib}
\usepackage{ulem}

\def\tsc#1{\csdef{#1}{\textsc{\lowercase{#1}}\xspace}}
\tsc{WGM}
\tsc{QE}
\tsc{EP}
\tsc{PMS}
\tsc{BEC}
\tsc{DE}

\begin{document}
\let\WriteBookmarks\relax
\def\floatpagepagefraction{1}
\def\textpagefraction{.001}

\shorttitle{Generation of synthetic delay time series for air transport applications}

\shortauthors{P Esteve et~al.}

\title [mode = title]{Generation of synthetic delay time series for air transport applications}

\author[1]{Pau Esteve}[]
\credit{Conceptualization of this study, Methodology, Software, Writing - original draft, Writing - review \& editing}

\author[1]{Massimiliano Zanin}[orcid=0000-0002-5839-0393]
\credit{Conceptualization of this study, Methodology, Software, Writing - original draft, Writing - review \& editing}

\affiliation[1]{organization={Instituto de F\'isica Interdisciplinar y Sistemas Complejos CSIC-UIB},
    addressline={Campus Universitat de les Illes Balears}, 
    city={Palma de Mallorca},
    postcode={E-07122}, 
    country={Spain}}

\cormark[1]

\ead{mzanin@ifisc.uib-csic.es}

\cortext[cor1]{Corresponding author}

\begin{abstract}
The generation of synthetic data is receiving increasing attention from the scientific community, thanks to its ability to solve problems like data scarcity and privacy, and is starting to find applications in air transport. We here tackle the problem of generating synthetic, yet realistic, time series of delays at airports, starting from large collections of operations in Europe and the US. We specifically compare three models, two of them based on state of the art Deep Learning algorithms, and one simplified Genetic Algorithm approach. We show how the latter can generate time series that are almost indistinguishable from real ones, while maintaining a high variability. We further validate the resulting time series in a problem of detecting delay propagations between airports. We finally make the synthetic data available to the scientific community.
\end{abstract}




\maketitle

\section{Introduction}

According to a quote of Daren Tang, Director General of the World Intellectual Property Organization (WIPO), {\it ``If digitalization is the engine of the future economy, then data is its fuel''}. The importance of data goes well beyond economy, to also directly impact science: nowadays it is difficult to find a discipline in which hypotheses are not, at least partly, generated and validated on large data sets - the only notable exception being pure theoretical fields, e.g. mathematics. Data are nevertheless not always easy to obtain: they may be limited in size, or their use may be constrained by confidentiality agreements. As a consequence, interest is growing around the idea of synthetic data: that is, data sets that are synthetically generated based on real ones and sharing their characteristics and usability \citep{emam2020chapter}.

Synthetic data can solve both the problem of scarcity and confidentiality. As they are generated, and not observed, any quantity can be created; they also do not disclose the starting information, at least if they are correctly generated \citep{stadler2020synthetic}, and hence do not reveal any private information. To illustrate a real application, synthetic data are expected to have a huge impact in medical education: models can be used to generate large quantities of training materials, that can be tuned (for instance to present the student with specific rare cases), but that are otherwise unique \citep{arora2020artificial, pataranutaporn2021ai}. Yet, training is only one of the potential applications. Data augmentation techniques allow to artificially expand data sets, thereby improving the performance of supervised learning tasks such as classification and forecasting \citep{iglesias2023data}. Another application of synthetic data is privacy preservation, enabling the creation of synthetic data sets that maintain the statistical properties of the original data without exposing confidential information \citep{jordon2018pate}.

While most data generation techniques have been developed in contexts like computer vision and natural language processing, some models have been adapted for time series generation. This may prima facie seem an easy task, as a time series can be seen as an image of size $n \times 1$, hence as an image of low dimensionality. Yet, time series data sets present some unique challenges. Image generation models are typically trained on massive collections containing millions \citep{deng2009imagenet} or even billions \citep{schuhmann2022laion} of images. In contrast, assembling similarly large data sets for time series is often impractical. For example, consider a univariate time series spanning 50 years of data. If we segment it into daily windows, we would obtain approximately 18,000 individual time series - a size far smaller than what is commonly used in computer vision. Moreover, time series data are inherently different from images: they exhibit strong temporal dependencies, are often imbalanced, and frequently come with privacy constraints. These factors make the adaptation of models from other fields to time series generation particularly challenging, adding significant complexity to the training of deep learning models in this domain compared to computer vision or natural language processing.

While the adoption of synthetic data sets in air transport has been lagging, several interesting solutions have been proposed in recent years. Most of them are based on the generation of synthetic trajectories, as for instance in Refs. \citep{park2021study, krauth2021synthetic, krauth2023deep, lukevs2023generating, gui2024novel, kanwal2024deep}.
Other examples include the generation of safety events \citep{miltner2014modeling, lalivs2018generating, aref2024generating, yesmin2025generative} and of flight networks \citep{fugenschuh2021structural}.

We here evaluate different options for generating time series of delays at airports, both for departure and landing. We specifically focus on the problem of generating realistic time series representing the evolution of the hourly average delay. This kind of aggregated data finds multiple applications in air transport. To illustrate, they are the basis of algorithms designed to unravel the propagation of delays between airports, through e.g. the concept of functional networks \citep{zanin2015can, zanin_network_2017, du_delay_2018, lpastorino_2022, jia_delay_2022}; and more generally, of models describing such propagations \citep{hansen2002micro, fleurquin2013systemic, pyrgiotis2013modelling, baspinar2016data, li2020data}. They can therefore be used to tune models, without the need to share the original data, while still retaining a high level of realism. While the data here generated do not account for couplings between airports, i.e. for actual propagations, they can be used as null models to exclude the appearance of false positives. These time series could also be used to generate disaggregated data, as e.g. delays of individual flights, by providing realistic indications of the expected delay at a given airport and time.

After introducing the real data supporting the generation (see Sec. \ref{sec:data}), we explore three algorithms; two of them based on established Deep Learning generative models (Sec. \ref{sec:DLmodels}), and a third one based on a simplified Genetic Algorithm approach preserving temporal correlations (Sec. \ref{sec:new_approach}). When compared using standard evaluation approaches and Deep Learning-based classification tasks, we obtain the surprising result that the latter model outperforms the first two; it further provides synthetic time series that are highly indistinguishable from real ones, while at the same time displaying a low correlation (i.e. they are not mere copies of the real data). In Sec. \ref{sec:data_anal} we further provide examples of how these data can be used, focusing on the problem of validating the reconstruction of functional connectivities, i.e. of delay propagation patterns. We finally make the full synthetic data set available to the scientific community, to foster future research on this topic (see Sec. \ref{sec:availability}).

\section{Real delay data}
\label{sec:data}

Data about the real hourly evolution of delays in airports have been extracted from two complementary sources. The first one is the EUROCONTROL's R\&D Data Archive, a public repository of European historical flights made available for research purposes and freely accessible at \url{https://www.eurocontrol.int/dashboard/rnd-data-archive}. The second one is the Reporting Carrier On-Time Performance database of the Bureau of Transportation Statistics, U.S. Department of Transportation, freely accessible at \url{https://www.transtats.bts.gov}.

Note that the two data sources have different temporal coverage: while EU's one includes only four months (i.e. March, June, September and December) of each year, the US data set includes all months; this limitation is defined at source, and cannot be avoided. Additionally, we have considered data starting in year 2015, being this the first year for which EUROCONTROL's data are available; and ending in year 2019, to avoid the disruption caused by COVID-19. This yields a total of $610$ and $1,825$ days, respectively for EU and US.

From these raw data, arrival (departure) delay time series have been extracted, calculated as the difference between the actual and scheduled landing (respectively, take-off) times of each flight, and averaged at each airport and at each hour of the day. Hence, for each airport and day, we generated two time series of length $24$. Data have further been limited to the top-$30$ airports in each region, according to the number of operations. No additional pre-processing of the data (e.g. normalisation or detrending) has been performed.

\section{Deep Learning approach}
\label{sec:DLmodels}

\subsection{State of the art in data generation using Deep Learning}
\label{sec:sota}

In recent years, various methods have been developed aimed at generating synthetic time series that preserve both the diversity and the statistical properties of the original data set. Most of the existing models in the literature can be grouped into two main families: Variational Autoencoders \citep{desai2021timevae, lee2023vector} and Generative Adversarial Networks \citep{esteban2017real, yoon2019time, pei2021towards, seyfi2022generating, wang2023aec}. On the one hand, VAEs encode input time series into a lower-dimensional latent space with an encoder, and then decode it to reconstruct the data, learning to generate new, similar time series by sampling from this latent space. In other words, VAEs try to compress the main characteristics of the time series, for then generating new ones respecting those patterns. On the other hand, GANs are generative models in which a generator learns to produce realistic time series by training to fool a discriminator, which dynamically evaluates the authenticity of the generated data. During training, the generator improves its output to deceive the discriminator, and once trained, it can generate new, realistic time series independently.

In addition to VAEs and GANs, flow-based models \citep{deng2020modeling, alaa2021generative} have also been proposed, which use invertible neural networks to directly model the distribution of the data. This is done through a sequence of invertible transformations that allow for both efficient sampling of new data and exact likelihood evaluation. Lastly, there are also mixed-type methods \citep{jeon2022gt, zhou2023deep} that combine elements of different model families.

The reader can find an extensive review of all these models in Ref. \citep{ang2023tsgbench}, together with a summary of publicly available data sets and evaluation metrics. Some of these models can be used with TSGM \citep{nikitin2023tsgm}, an open-source Python library for synthetic time series generation. These techniques have found applications across diverse fields, including healthcare \citep{esteban2017real, yan2022multifaceted, li2023causal}, finance \citep{dogariu2022generation, ramzan2024generative}, and telecommunications \citep{lin2020using}, where access to large, high-quality data sets is often restricted.

\subsection{Time series generation}

Among the various possible models, we here consider a Variational Autoencoder (TimeVAE) and a Generative Adversarial Network (TimeGAN). These models were chosen because they represent the two main families of deep generative models and are widely used as benchmark baselines in the literature \citep{ang2023tsgbench}. Additionally, they are not domain-specific and have been applied in a variety of different fields, as illustrated above.

\begin{figure}
    \centering
    \includegraphics[width=1.00\linewidth]{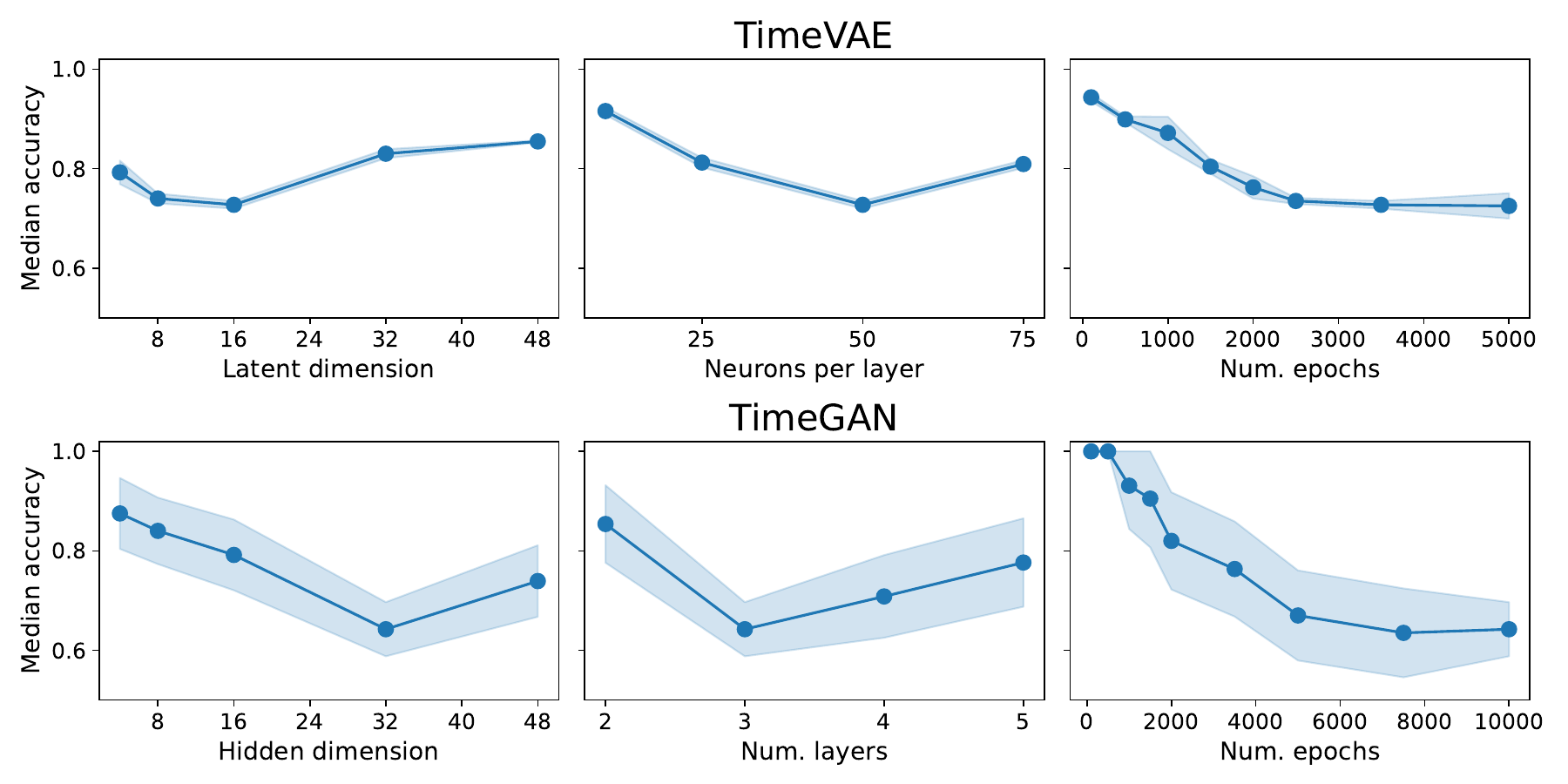}
    \caption{Hyperparameter evaluation for TimeVAE (top) and TimeGAN (bottom). For each combination, 20 synthetic data sets for London Heathrow were generated. Lines show the median discriminative score, and shaded bands indicate the standard deviation. Each subpanel explores a range of values for the corresponding hyperparameter, keeping the others at their optimal values. For TimeVAE, the architecture is limited to three layers of sizes $n \times (1, 2, 4)$, where $n$ is the number of neurons per layer.}
    \label{fig:hyperparametersDL}
\end{figure}

\begin{itemize}
    \item \textbf{TimeVAE} \citep{desai2021timevae}. As previously discussed, VAEs are generative models that use an encoder to map data into a lower-dimensional latent space, and a decoder to generate new data samples from this latent representation. TimeVAE extends VAEs to multivariate time-series generation. In order to find the best hyperparameters for this model, we have performed several analyses varying them and using the time series for London Heathrow (EGLL) as a reference. The top panels of Fig. \ref{fig:hyperparametersDL} then report the obtained median accuracy in a task of discriminating real from synthetic time series, based on a ResNet model - details will be discussed in Sec. \ref{sec:dl_discr}. The best results (i.e. lower accuracy) were obtained for both an encoder and a decoder with three hidden layers of sizes 50, 100, and 200 each, and a latent dimension of 16.
    \item \textbf{TimeGAN} \citep{yoon2019time}. GANs are generative models in which a generator learns to produce realistic data by indirectly training to fool a discriminator, which dynamically evaluates the authenticity of the generated data. TimeGAN extends GANs to multivariate time-series generation by optimising a joint embedding space with both adversarial and supervised objectives. As in the TimeVAE case, we use the optimal hyperparameters found in Fig. \ref{fig:hyperparametersDL}, see bottom panels, including a generator and a discriminator both with three hidden layers, a hidden dimension of 32 and a design based on Gated Recurrent Units \citep{cho2014learning}.
\end{itemize}

Each model takes as input the European data set, with time series representing the average delay per hour for 30 different airports. This data set contains 610 time series, each of length 24, as described in Sec. \ref{sec:data}. All time series are used in the training process, obtaining an output of the same size. To account for the stochastic nature of the generative models, we repeat the data generation process 20 times, with independent realisations. Thus, for each input, we generate 20 independent synthetic data sets of shape $610 \times 24$. This approach is particularly important for TimeGAN, which exhibits inherent variability, as reflected in the wider shaded bands in Fig. \ref{fig:hyperparametersDL}.

\subsection{Evaluation using dimensionality reduction}

Once the synthetic data have been generated, it is important to validate them, that is, to see how similar they are to the original ones. We here tackle this point by initially resorting to the most common solution in the literature, i.e. the graphical analysis of dimensional-reduced data. Dimensionality reduction techniques project both the original and synthetic data sets in a low-dimensional space, allowing for a visual comparison of their respective distributions. In simple terms, each time series of $24$ values is represented by a point in a plane, such that two points near in space represent two time series of high similarity. When comparing the position of points of real and synthetic time series, if these occupy the same portion of the plane, it can be concluded that they are similar.

\begin{figure}
    \centering
    \includegraphics[width=1.00\linewidth]{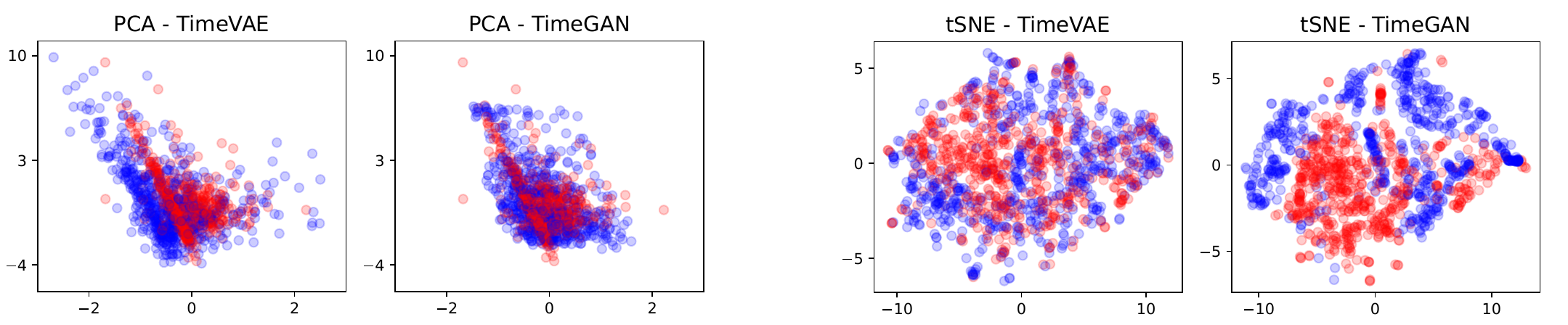}
    \caption{PCA and tSNE visualisations comparing original (red) and synthetic (blue) data sets generated using TimeVAE and TimeGAN, as indicated in the title of the respective figures. Data correspond to arrival delay time series for London Heathrow (EGLL) airport.}
    \label{fig:visualTest_DL}
\end{figure}

In Fig. \ref{fig:visualTest_DL}, we project both data sets into a 2D space using both linear (PCA, left panel, \citep{greenacre2022principal}) and non-linear (tSNE, right panel, \citep{van2008visualizing}) dimensionality reduction methods. While slight differences between the original (red) and synthetic (blue) data sets can already be observed with PCA, these are much more noticeable with tSNE. It can be concluded that neither TimeVAE nor TimeGAN manages to capture the distribution of the original data set, corresponding to delay time series at London Heathrow airport.

\subsection{Evaluation using discriminative score}
\label{sec:dl_discr}

While widespread in the literature, the previous dimensionality reduction approach can only convey qualitative information about the similarity between two data sets. 
Especially when aiming at real applications, it is essential to quantify the fidelity between the original and synthetic data sets beyond a simple visual comparison, as done in the previous section. A common approach is to employ a discriminative score \citep{ang2023tsgbench}: a post-hoc neural network is trained to distinguish between real and synthetic time series in a supervised binary classification task. Each time series is labelled as either real or synthetic, and the discriminator is trained to classify them accordingly. The resulting discriminative score measures how distinguishable the synthetic time series are from the originals, with lower values indicating greater similarity; to illustrate, a score of $0.50$ corresponds to synthetic data that are indistinguishable from the original. 

In our case, we use a Residual Network (ResNet) model, i.e. artificial neural networks inspired by the way pyramidal cells are organised in the cerebral cortex; specifically, the connections between layers are not sequential, but instead some layers can be skipped (creating shortcuts or jumps). This presents the advantage of a simpler structure, and consequently of a reduced training cost \citep{he2016deep}. The configuration here considered consists of two blocks and three layers per block, with a number of epochs in training equal to $50$; these values have previously been demonstrated to be effective in similar classification tasks \citep{crespo2024deep}. Each classification task is performed on a single airport, using a random half of the real data and a random half of the synthetic data for training; the task is evaluated on the remainder of the data using the accuracy, i.e. the fraction of time series correctly classified. In order to account for the natural stochasticity of the training and evaluation, the final score corresponds to the median of the accuracy over $50$ independent realisations.

While synthetic time series ought to be similar to the original ones, it is also important to ensure that they are not the same, i.e. that some variability is retained. One can easily imagine a generative algorithm creating copies of the original time series; such approach would yield very low discriminative scores, as the discriminator would be unable to distinguish them from the originals; yet, the synthetic time series would be useless. To assess this, we compute the correlation score, which measures the similarity between each synthetic time series and all original time series, retaining the maximum correlation value. In other words, a value of one would indicate that the synthetic time series is a mere copy of one in the original set; on the other hand, the lower the value, the higher is the novelty introduced. By repeating this process for each time series, we obtain a distribution of correlation values for the whole synthetic data set. A high-quality synthetic data set should achieve both a low discriminative score and a low correlation score.

\begin{figure}
    \centering
    \includegraphics[width=1.0\linewidth]{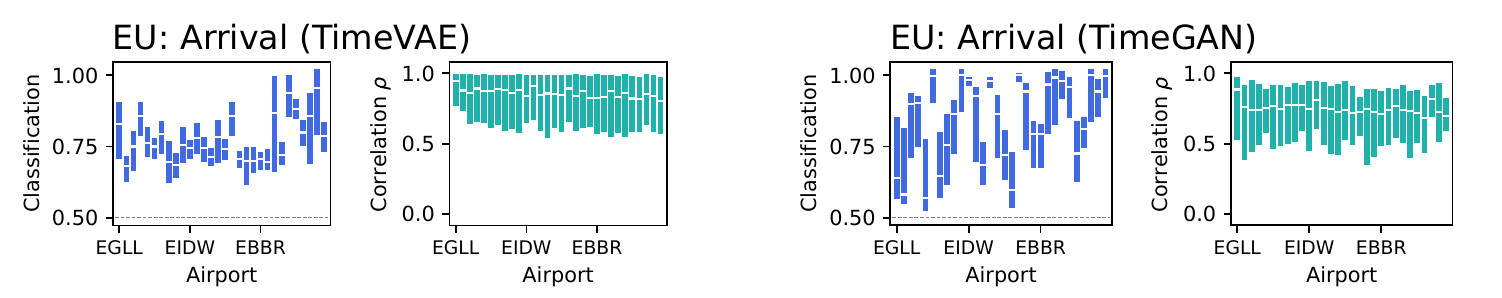}
    \caption{Distribution of classification scores (left panels, blue bars) and of correlations (right panels, cyan bars) for the synthetic data generated with TimeVAE \citep{desai2021timevae} (left) and TimeGAN \citep{yoon2019time} (right) for the case of arrival delay time series in Europe. Each bar represents the range between the maximum and minimum of the distribution; the horizontal white line reports the median. See main text for definitions.}
    \label{fig:global_score_DLmodels}
\end{figure}

Fig. \ref{fig:global_score_DLmodels} presents an analysis of both the discriminative and correlation scores for synthetic data generated using TimeVAE and TimeGAN. In the case of TimeVAE, the median discriminative score consistently exceeds $0.70$, with some airports exhibiting particularly high values, implying that the synthetic data are easily distinguishable from the real ones. For TimeGAN, the values span a wider range, yielding satisfactory results for only a limited number of airports. However, in both cases, the synthetic data exhibit a high correlation with the original data set, indicating that the models generate highly similar time series rather than truly novel samples. This effect is particularly pronounced in TimeVAE, where the encoding and subsequent decoding of each time series result in outputs that are just slightly modified versions of the original input.

\section{Simplified Genetic Algorithm approach}
\label{sec:new_approach}

As an alternative to the previously explored approaches based on Deep Learning, we here consider a simpler algorithm incorporating basic information about the evolution of delays in air transport. It is based on two elements, which are tackled in what follows: the creation of individual vectors of daily delays, on the one hand; and the consolidation and refinement of such vectors into a single data set for each airport, on the other hand. As will be detailed, this is similar to a simplified Genetic Algorithm approach \citep{holland1992genetic, mitchell1998introduction, kramer2017genetic}, in that time series are retained according to a fitness score; yet, the generation is simplified, and based on known aspects of air traffic dynamics. These two elements are respectively discussed in Secs. \ref{sec:one_vector} and \ref{sec:one_dataset}; an initial evaluation of the results is further reported in Sec. \ref{sec:eval_new}.

\subsection{Creation of one delay vector}
\label{sec:one_vector}

This first part of the algorithm aims at creating a single vector $d(t)$ of $24$ elements, hence with $t \in [1, \ldots, 24]$, representing the average hourly delay (either at departure or arrival) for a given airport. For the sake of clarity, below we refer to synthetic values as $d(t)$, and to the set of real values at time $t$ and across all days as $d^r(t)$.

Defining the values of the first elements of such vectors is a challenging task, as these correspond to night hours. Firstly, few operations are expected at that time, hence the average delay can strongly fluctuate. Secondly, the delay of those operations will depend on what happened the day before; yet, synthetic delays are constructed one day at a time, hence inter-day propagation is not considered. In order to solve this, we have opted for a simple random sampling of the average delay observed at the same airport and at the same time, across all available days. Note that such random sampling ensures that the value $d(t)$ will be realistic, as indeed it will correspond to a value historically observed in the system; at the same time, no correlations between consecutive values, i.e. between $d(t)$ and $d(t+1)$, will be present. As previously described, this random sampling is only applied to those hours at the beginning of each day with few operations; after a manual analysis of the data, we have opted for applying it only to the first four values of the vector $d$. 

We then move to the definition of the remainder elements $d(t)$ with $t > 4$. While a possible solution could be to continue with the random sampling used in the first part (as will be discussed in Sec. \ref{sec:eval_new}), this would come at the important cost of neglecting the correlations between consecutive values. Yet, we know these are important. Firstly, from an operational perspective, it is to be expected that operations scheduled to land/depart in peak hours may have to be moved to subsequent hours due to limited capacity; hence, delays at one hour will depend on the previous ones, especially in periods of high delays. Secondly, it has previously been shown that delay vectors of different airports are highly identifiable by Deep Learning classification models \citep{ivanoska2022assessing, gil2024low}, which are mostly based on convolutional operations; or, in other words, which especially detect correlations between consecutive values.

In order to account for such correlation, we start by looking at the real values for the previous hour $d^r(t-1)$, and divide all available values into ten deciles. Next, we observe in which one of these deciles the synthetic value previously generated for $t-1$ falls; and obtain the distribution of the next value $d^r(t)$ for those $d^r(t-1)$ inside that decile. Note that this allows us to calculate a distribution of the expected values of delays at time $t$, given how delays have actually evolved when the observed delay at time $t-1$ was similar to what synthetically generated. As a final step, we divide the obtained distribution into its ten deciles; select one of them at random; and generate the new synthetic value $d(t)$ as a random number uniformly distributed between the limits of that decile. In short, this allows to retain correlations between consecutive values, as we observe how their dynamics evolved given the value at time $t-1$; yet, generated values are not mere copies, but rather extracted from distributions that are similar to the original ones.

For the sake of clarity, below we synthesise the main steps of this part of the algorithm, i.e. to generate values when $t > 4$:

\begin{enumerate}
    \item Take the real values observed in the previous hour, i.e. all $d^r(t-1)$ across all days, and divide them into 10 deciles. 
    \item Evaluate in which decile the synthetic value $d(t-1)$ falls.
    \item Select the vectors whose value $d(t-1)$ is inside that decile, and obtain their distribution for the next time step $d^r(t)$.
    \item Divide the previous distribution again in 10 deciles, and randomly select one of them.
    \item Generate a random number uniformly distributed between the limits of that decile.
\end{enumerate}

\subsection{Creation of one delay data set}
\label{sec:one_dataset}

As discussed in Sec. \ref{sec:data}, the original delay data for a single airport are organised in matrices of size $610 \times 24$ for Europe and $1,825 \times 24$ for the US, where $610$ and $1,825$ are the number of available days. It thus makes sense to generate synthetic data sets of the same size, by merging the same number of delay vectors, each one independently generated using the algorithm described above. Yet, due to the stochastic nature of the generation, some of these vectors may be indistinguishable from the real ones, while others may easily be spotted; a further refinement is therefore required.

Given one set of synthetic delay vectors, a random half of these and a random half of the real delay data are combined to train a ResNet classification model, whose objective is to discriminate which ones of these vectors are synthetic and which ones are real. This ResNet model shares the same architecture as the one described in Sec. \ref{sec:dl_discr}; the number of epochs in the training has nevertheless been reduced to $20$. This was done to reduce the computational cost; at the same time, obtaining perfect classifications is here not required. This model is then applied to the remainder part of the synthetic vectors; those that are correctly identified as synthetic by the model are discarded, and substituted with new synthetic delay vectors. In other words, we leverage a simplified Deep Learning model to single out those synthetic vectors that are easy to be spotted, and to substitute them with new ones. The whole process is repeated $10^3$ times, to obtain the final synthetic delay data set.

\subsection{Performance evaluation}
\label{sec:eval_new}

\begin{figure}
    \centering
    \includegraphics[width=0.50\linewidth]{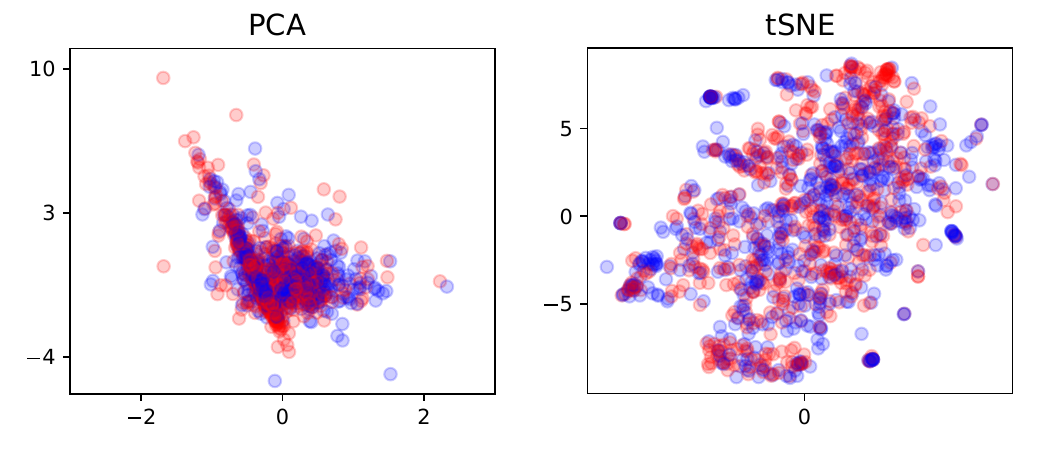}
    \caption{PCA and tSNE visualisations comparing original (red) and synthetic (blue) data sets generated using the simplified Genetic Algorithm approach. Data correspond to arrival delay time series for a single airport, London Heathrow (EGLL).}
    \label{fig:visualTest_OurMethod}
\end{figure}

As done with the other generative models in Sec. \ref{sec:DLmodels}, we here perform a first evaluation of the performance of the generation model. We start with plots of the dimensionality-reduced data, i.e. akin to those in Fig. \ref{fig:visualTest_DL}; as opposed to the previous models, here original (red) and synthetic (blue) time series seem to strongly overlap in the plane, see Fig. \ref{fig:visualTest_OurMethod}.

\begin{figure}
    \centering
    \includegraphics[width=1.0\linewidth]{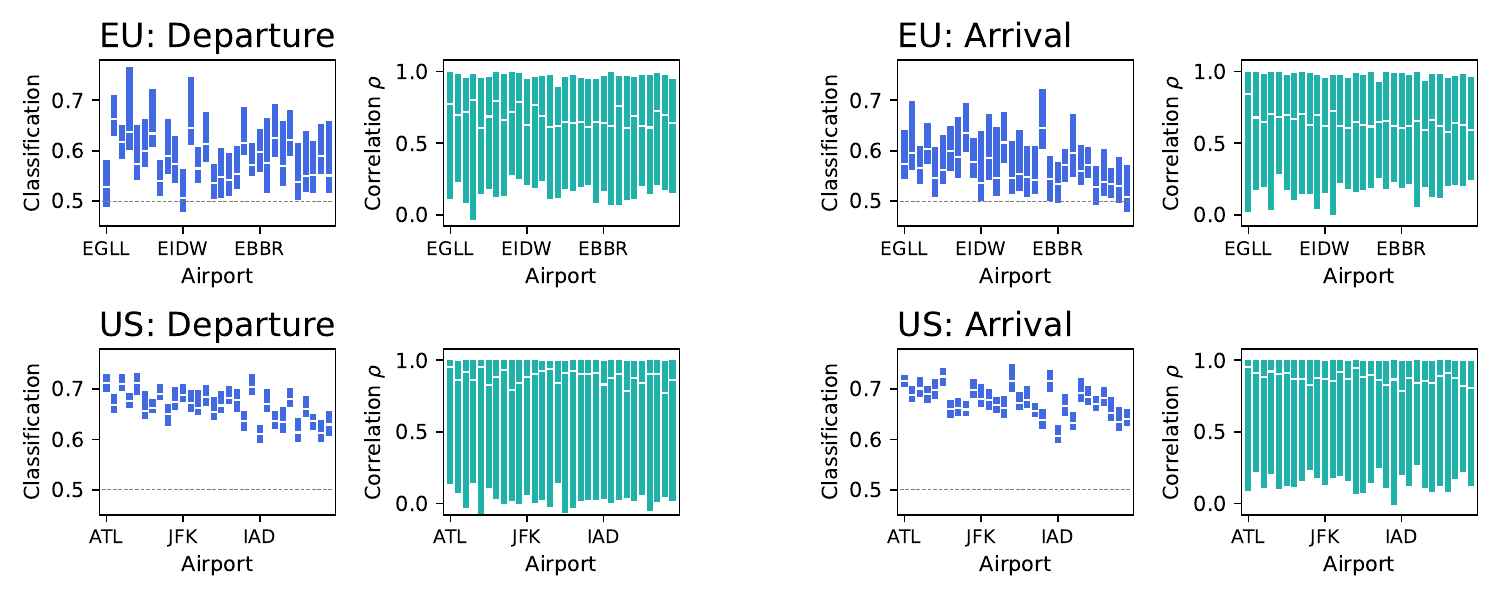}
    \caption{Distribution of classification scores (left panels, blue bars) and of correlations (right panels, cyan bars), for the considered airports in Europe (top row) and US (bottom row), and for departure (left column) and arrival (right column) delays. Each bar represents the range between the maximum and minimum of the distribution; the horizontal white line reports the median. See main text for definitions. Full results are further reported in the Appendix.}
    \label{fig:global_score}
\end{figure}

Fig. \ref{fig:global_score} reports a synthesis of the classification score obtained by a ResNet model trained to discriminate between synthetic and real delays, as this has proven to be the strictest test. Specifically, each blue bar in the left panels of Fig. \ref{fig:global_score} reports the minimum and maximum classification score for each airport over $100$ independently created synthetic data sets (i.e., when the process described in Sec. \ref{sec:one_dataset} is repeated $100$ times); the white line in the middle indicates the corresponding median. It can be seen that results are globally good, with median classification scores generally lower than $0.6$ for Europe, and than $0.7$ in the case of the US; in other words, the ResNet models find extremely challenging to identify synthetic delays. As in Sec. \ref{sec:dl_discr}, we calculated the correlation between synthetic and real delays, see cyan bars in the right panels of Fig. \ref{fig:global_score}. It can be appreciated that the distributions of such correlations are very wide, spreading from almost zero to one. Still, the medians for Europe oscillate around 0.7, thus suggesting that synthetic delays are not simple copies of the real data.

\begin{figure}
    \centering
    \includegraphics[width=0.6\linewidth]{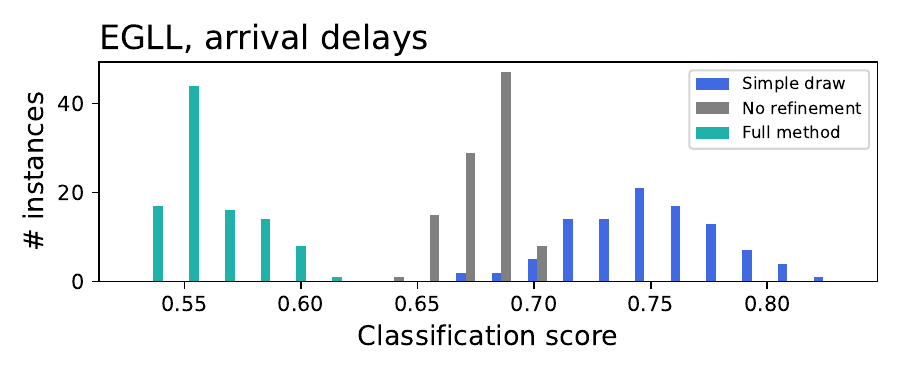}
    \caption{Probability distributions of the classification scores corresponding to arrival delays for London Heathrow (EGLL), as measured in the synthetic data generated using the full algorithm (cyan bars); using vectors generated by simple random drawn (blue bars); and by omitting the refinement process of Sec. \ref{sec:one_dataset} (grey bars).}
    \label{fig:random_draw}
\end{figure}

Before moving to the full analysis of the generated synthetic data, we discuss two alternatives that were previously hinted: the generation of synthetic delays by simply drawing real values at random, see Sec. \ref{sec:one_vector}; and avoiding the use of the refinement procedure described in Sec. \ref{sec:one_dataset}. For one single case corresponding to arrival delays at London Heathrow, Fig. \ref{fig:random_draw} reports the histograms of the probability distribution of the classification scores for the full method (cyan bars); for the full method, but drawing delay values at random (dark blue bars); and for data sets created by merging synthetic vectors, but without any posterior refinement (grey bars). It can be appreciated that the generation of delay vectors using simple random data, i.e. not taking into account correlations between consecutive values, yields results that are halfway between the best solution and what obtained in Sec. \ref{sec:DLmodels}. Similar results, albeit slightly better, are also obtained when omitting the refinement. These two variants, and especially omitting the refinement process, could therefore be a solution whenever the computational cost is a concern, in exchange for less realistic results.

\begin{figure}
    \centering
    \includegraphics[width=1.0\linewidth]{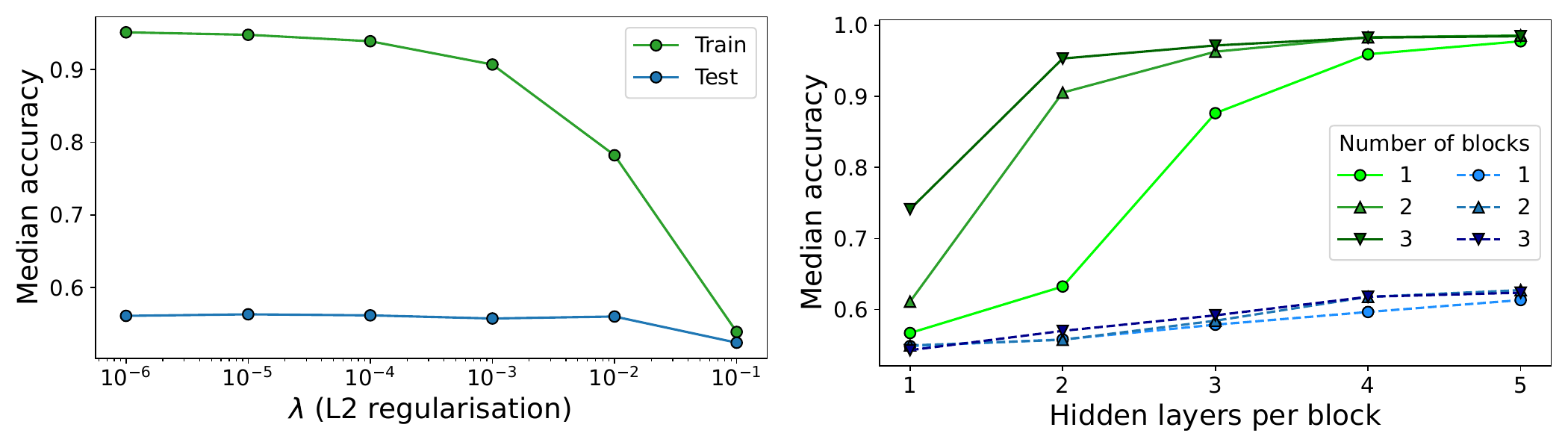}
    \caption{Hyperparameter analysis of the ResNet models. (Left) Median classification accuracy as a function of the L2 regularisation applied to all convolutional layers. (Right) Median classification accuracy as a function of architectural structure, i.e. the number of blocks (symbol shapes, see legend) and hidden layers per block. Green and blue correspond to the training and testing sets, respectively.}
    \label{fig:hyperparameters}
\end{figure}

As a final point, we want to exclude the possibility that the low classification scores obtained when comparing synthetic and real delays may be caused by sub-optimal hyperparameter choices in the ResNet model. Fig. \ref{fig:hyperparameters} reports the evolution of the median classification score, across all European airports for arrival delays, as a function of the rate of the L2 regularisation (left panel) and of the number of hidden layers per block (right panel). It can be appreciated that the accuracy for the test sets does not substantially vary, as opposed to the one for the train sets. The major difference between them suggests that the models are overfitting; still, compensating for this, for instance by increasing the regularisation rate, does not improve the overall results.

\section{Synthetic data analysis}
\label{sec:data_anal}

The synthetic data generated in Sec. \ref{sec:new_approach} seem promising, at least when evaluated using a classification task; in this section we are going to confirm this, by performing additional tests inspired by the use cases sketched in the introduction.

\begin{figure}
    \centering
    \includegraphics[width=1.0\linewidth]{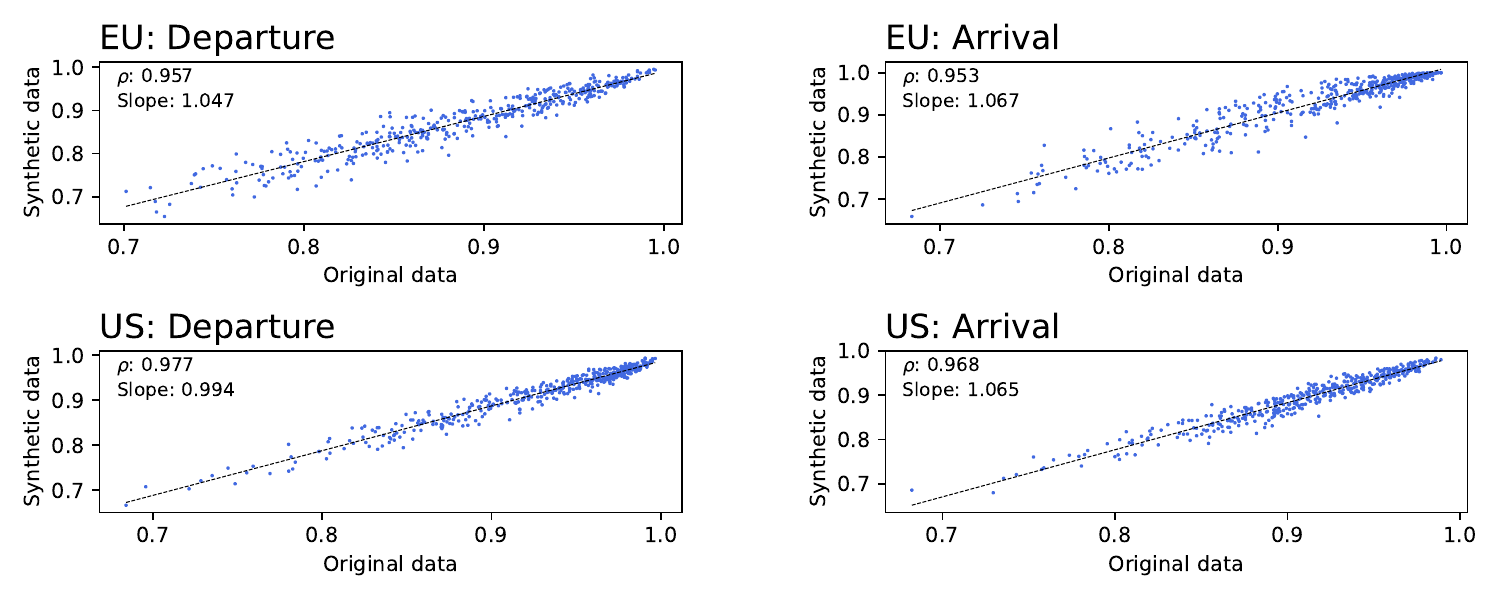}
    \caption{Classification score obtained by ResNet models when discriminating pairs of airports. Each point reports the accuracy in the classification of two airports obtained using the synthetic data (Y axes), as a function of the score for the real data (X axes). The four panels correspond to Europe (top row) and US (bottom row), and to departure (left column) and arrival (right column) delays.}
    \label{fig:cross_class}
\end{figure}

Firstly, we consider the results presented in Ref. \citep{ivanoska2022assessing}, according to which daily delay profiles of different airports are highly identifiable; in other words, given two airports, a model can be trained to recognise which one of them a vector of delays corresponds to. If delay profiles of different airports have unique characteristics, such characteristics should be preserved in the corresponding synthetic data, and airports should therefore also be identifiable using the latter ones. Fig. \ref{fig:cross_class} then reports scatter plots of the classification score when using the real (X axes) and synthetic data (Y axes). Each point corresponds to the accuracy between a pair of airports, obtained using ResNet models (same architecture as before). It can be appreciated that results are almost perfectly correlated; synthetic data thus retain the uniqueness of each airport.

\begin{figure}
    \centering
    \includegraphics[width=1.0\linewidth]{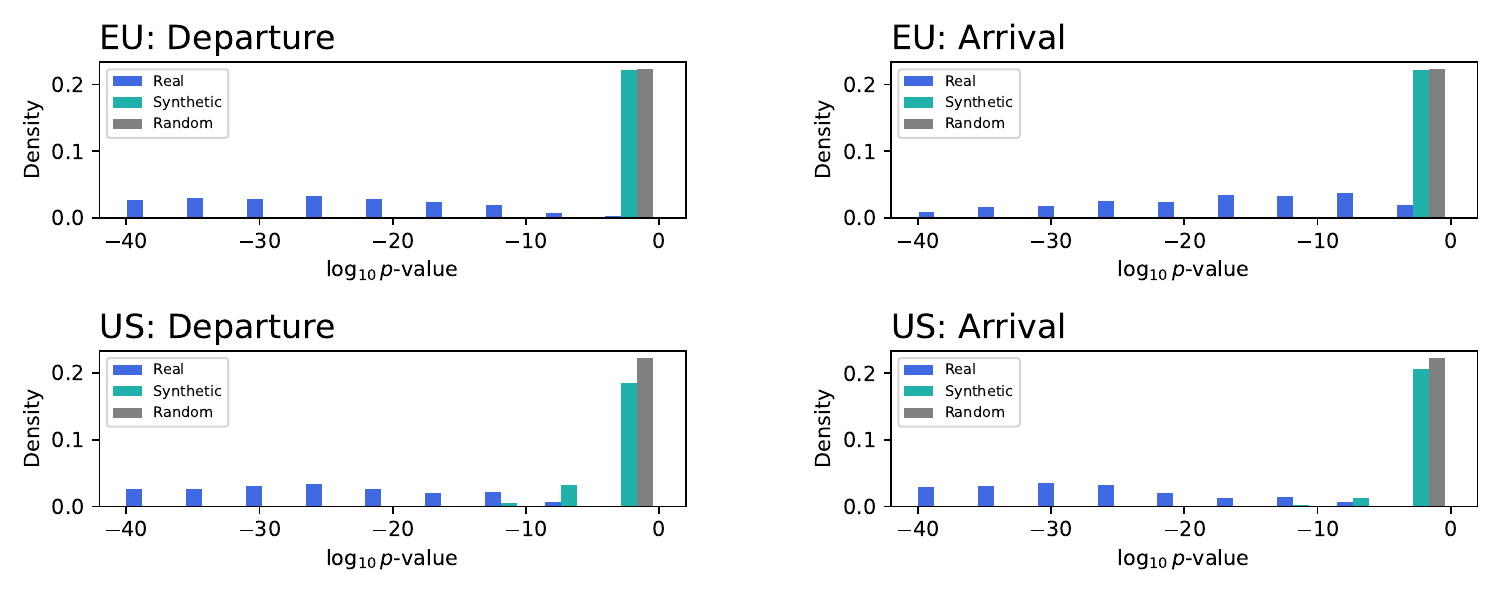}
    \caption{Histograms of the $\log_{10}$ of the $p$-value obtained by a GC test between pairs of airports, using: the real time series (blue bars); the synthetic data (cyan bars); and randomly shuffled version of the real time series (grey bars). The four panels correspond to Europe (top row) and US (bottom row), and to departure (left column) and arrival (right column) delays.}
    \label{fig:granger}
\end{figure}

Secondly, we move to the problem of detecting the propagation of delays, using the approach based on reconstructing functional networks \citep{zanin2015can, zanin_network_2017, du_delay_2018, lpastorino_2022, jia_delay_2022}. Note that, in this case, we expect results to be different in the real and synthetic data - as the latter ones were created one airport at the time, and therefore cannot include any propagation. Given the time series for two airports, we test the presence of propagation using the well-known Granger Causality (GC) test. This test assesses the presence of a ``predictive causality'' between two time series, i.e. instances in which the past of one of them helps predicting the future evolution of a second \citep{granger1969investigating}. Additional details on the methodology can be found in several papers, e.g. in Ref. \citep{lpastorino_2022}.
Fig. \ref{fig:granger} reports histograms of the $p$-values yielded by the GC test, when applied to the time series of pairs of airports. When we compare what obtained for the real (blue bars) and the synthetic data (cyan bars), it can be appreciated that only for the former the obtained $p$-values are small enough to be statistically significant. On the contrary, what obtained for the synthetic data is almost equal to the results corresponding to randomly shuffling the real data (grey bars), hence destroying any temporal structure. In short, and as expected, synthetic delays preserve the characteristics at each airport, but not the propagation between pairs of them.

\section{Data set availability and structure}
\label{sec:availability}

The synthetic time series generated in this work are freely available at \url{https://doi.org/10.5281/zenodo.15046397}, and are organised in four files: {\it EUArr.npy}, {\it EUDep.npy}, {\it USArr.npy}, and {\it USDep.npy}. Each one of them is a standard NumPy array, that can be opened with the corresponding Python library \citep{van2011numpy, idris2015numpy, harris2020array}. Each file contains a single four-dimensional tensor of size $30 \times 100 \times d \times 24$, the four dimensions respectively representing (i) the $30$ airports of the region, (ii) the $100$ independent realisations of the generation process, (iii) the $d$ days available for each region ($610$ for EU and $1,825$ for US), and (iv) the $24$ hours of the day. For the European data sets, each value represents the average hourly delay of the corresponding airport in seconds, while for the US the values are expressed in minutes, following the content of the corresponding source. The full list of airports for both regions is included as text files.

\section{Discussion and conclusions}

The use of synthetic data sets is a topic gaining momentum in air transport and air traffic studies. While solutions have been proposed to generate e.g. synthetic trajectories, to the best of our knowledge, no generative models have been applied to more macro-scale data. We here tackled the problem of synthesising time series representing average departure and arrival delays at major European and US airports. Three models are compared, two of them based on established Deep Learning architectures, and one on a simplified Genetic Algorithm approach. Validations, both qualitative and quantitative, confirm that the synthetic data are highly similar to the original ones, while still displaying a high variability, i.e. they are not mere copies (see Fig. \ref{fig:global_score}). They can also be used to reproduce or validate previous results obtained in the literature (see Sec. \ref{sec:data_anal}). We finally made public these synthetic data, to foster both applications and further research on the topic (Sec. \ref{sec:availability}).

While promising, the results here presented highlight some challenges that will have to be tackled in the future. Firstly, the analysed time series (and hence, those generated) have a low granularity, as they represent the average dynamics in one-hour intervals. Applications may nevertheless benefit from higher temporal resolutions; in the limit, one may want time series representing the delay of each individual operation. This nevertheless entails a higher complexity, due to the increased dimensionality of the time series to be generated, and therefore to the increased computational cost and real data requirements. A solution may come from the use of the proposed time series as starting point: given the expected average delay at a given hour, the delay of individual operations therein can then be synthesised according to some stochastic rule.

Secondly, the main validation of the generated data was here performed using a ResNet classification model. It is important to note that the detected similarity is a function of the sensitivity of the considered model. In other words, if a simpler model were used, the time series described in Sec. \ref{sec:DLmodels} may have passed such test; conversely, a more powerful model may detect differences in the synthetic time series of Sec. \ref{sec:new_approach}. Similarly, there may be applications where the time series presented in Sec. \ref{sec:DLmodels} may be good enough, while other ones may have more strict requirements. In short, the validation of synthetic data is not a closed or immutable task: it is rather dependent on the available technology and on the specific application.

As a last point, it is interesting to highlight that the problem here tackle is not a mere computational exercise, but can also help understanding the idiosyncrasies of individual airports. Specifically, it can be appreciated in Figs. \ref{fig:global_score_DLmodels} and \ref{fig:global_score} that the classification score strongly varies between airports. Whenever an airport is characterised by a clear pattern, the synthesis of its time series is expected to be an easy task; the same holds true if its delays were completely random. The results here reported thus hint at some airports having complex delay patterns, not easily captured by the three synthesis models here considered, and thus resulting in higher classification scores. The analysis and characterisation of these patterns may represent an interesting research venue.

\newpage
\clearpage

\appendix
\section{Appendix}

\begin{figure}[!h]
    \centering
    \includegraphics[width=1.0\linewidth]{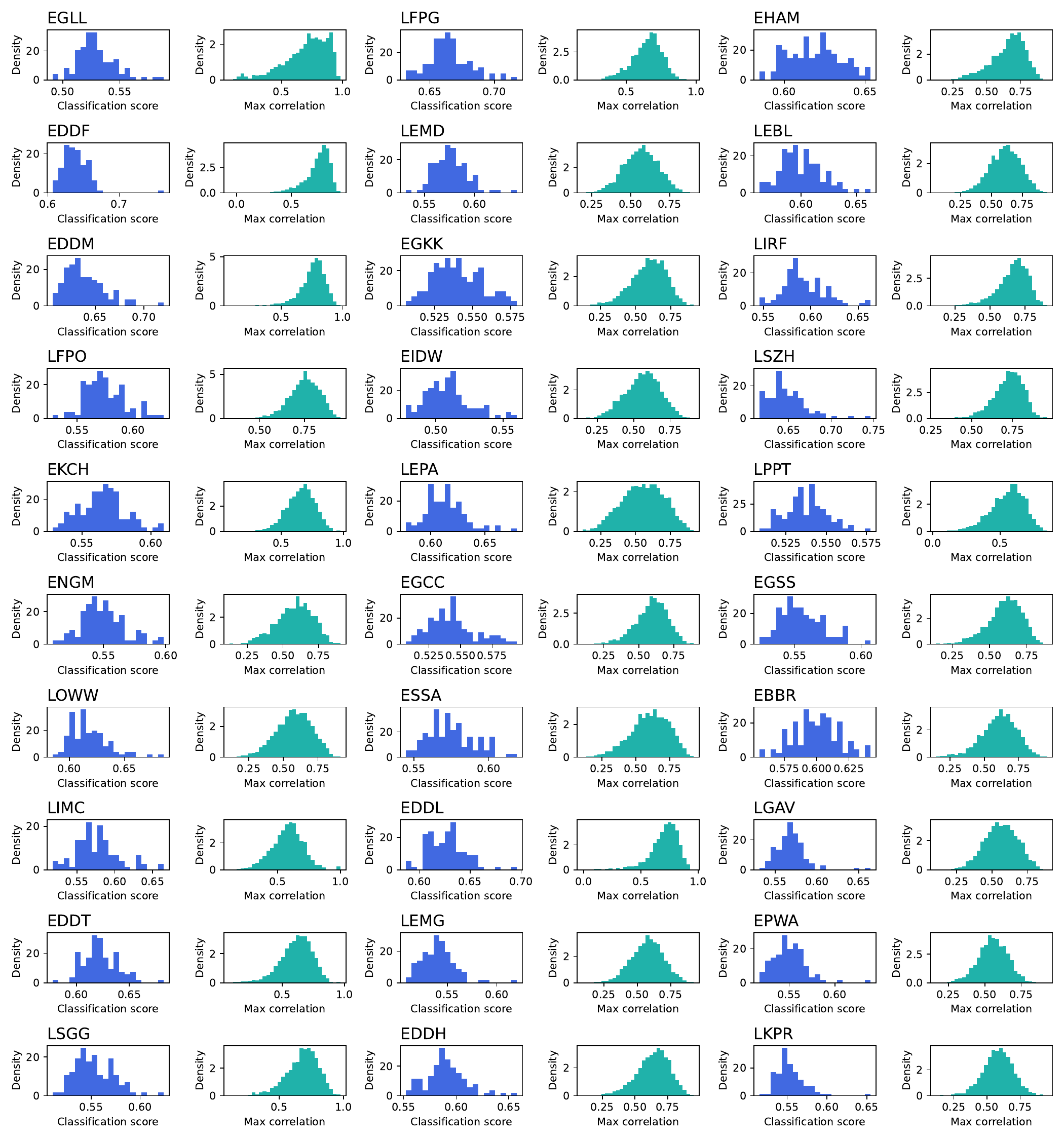}
    \caption{Histograms of the classification scores (left panels, blue bars) and of the correlations (right panels, cyan bars) for synthetic time series: European airports and departure operations. }
    \label{fig:aux_EUDep}
\end{figure}

\begin{figure}[!h]
    \centering
    \includegraphics[width=1.0\linewidth]{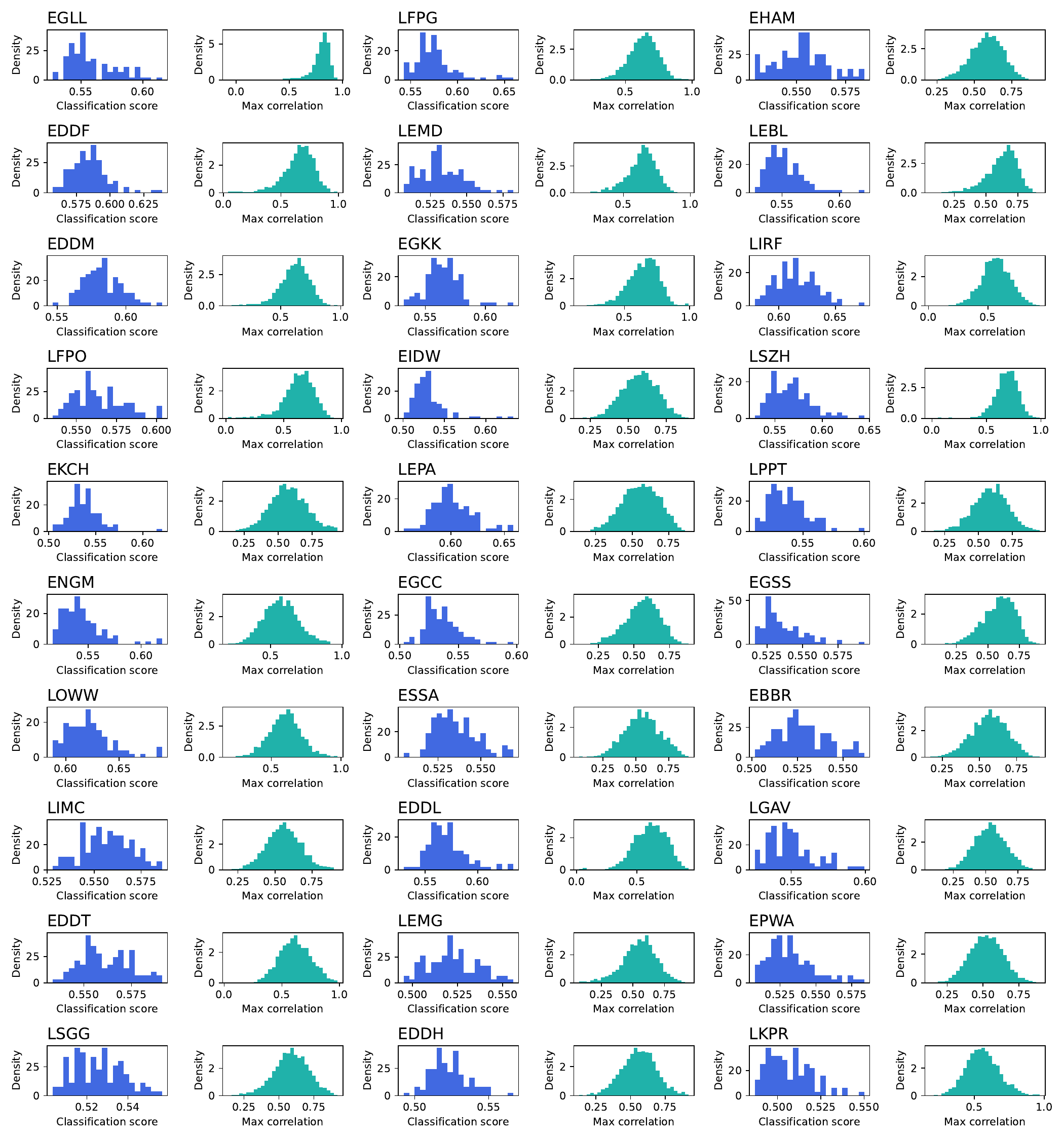}
    \caption{Histograms of the classification scores (left panels, blue bars) and of the correlations (right panels, cyan bars) for synthetic time series: European airports and arrival operations.}
    \label{fig:aux_EUArr}
\end{figure}

\begin{figure}[!h]
    \centering
    \includegraphics[width=1.0\linewidth]{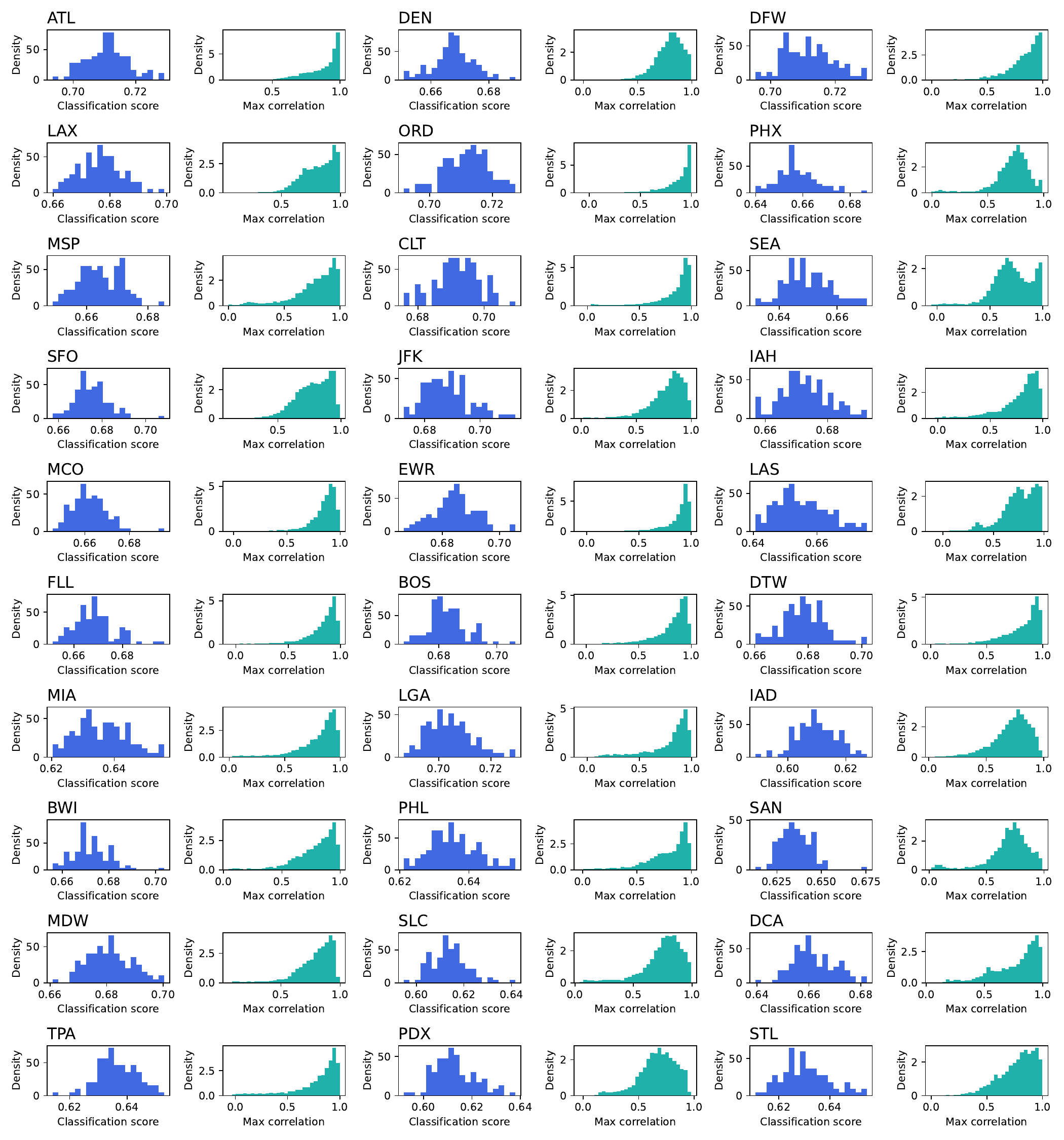}
    \caption{Histograms of the classification scores (left panels, blue bars) and of the correlations (right panels, cyan bars) for synthetic time series: US airports and departure operations.}
    \label{fig:aux_USDep}
\end{figure}

\begin{figure}[!h]
    \centering
    \includegraphics[width=1.0\linewidth]{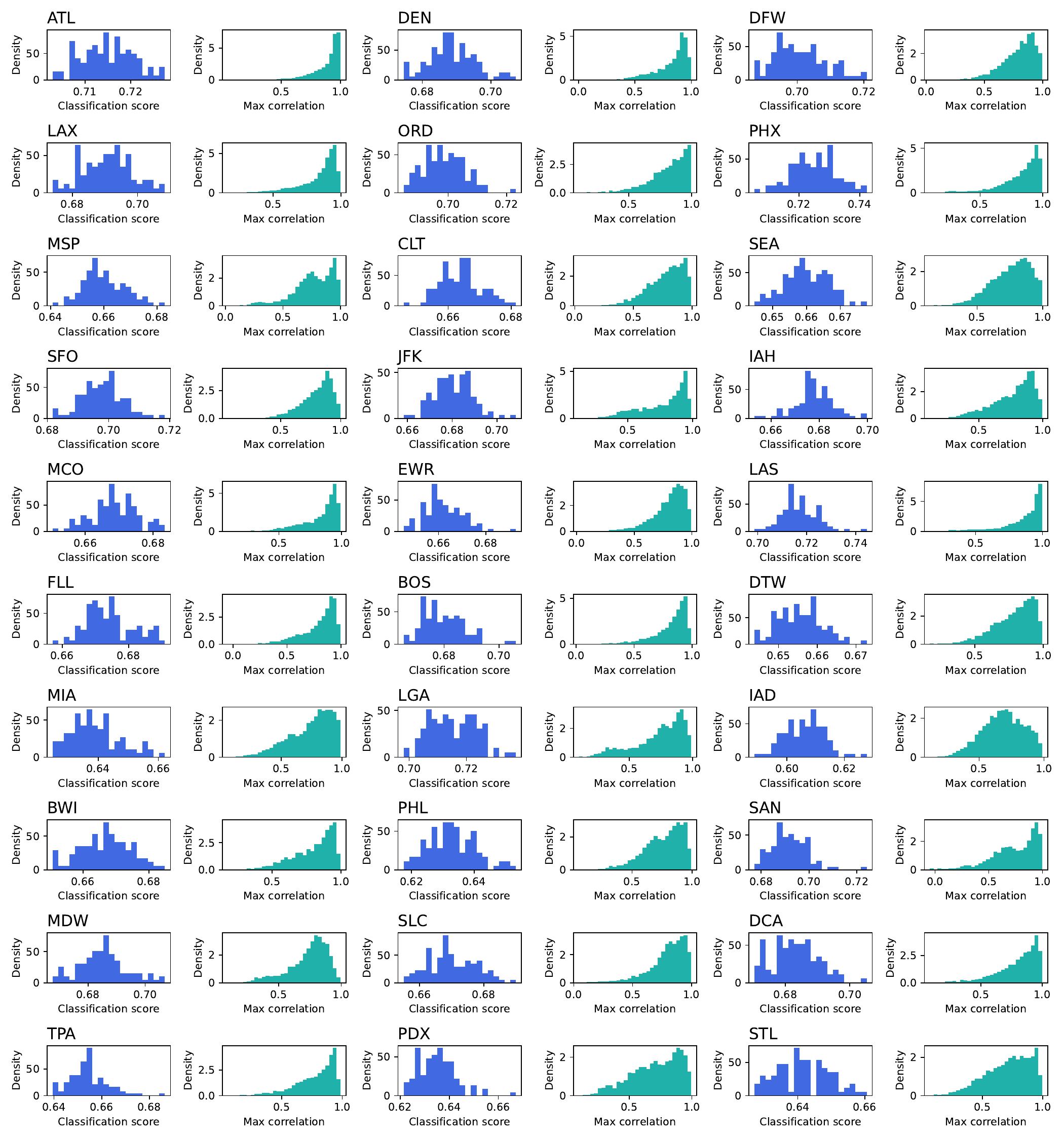}
    \caption{Histograms of the classification scores (left panels, blue bars) and of the correlations (right panels, cyan bars) for synthetic time series: US airports and arrival operations.}
    \label{fig:aux_USArr}
\end{figure}

\newpage
\clearpage

\section*{Acknowledgements}

This project has received funding from the European Research Council (ERC) under the European Union's Horizon 2020 research and innovation programme (grant agreement No 851255). This work was partially supported by the Mar\'ia de Maeztu project CEX2021-001164-M funded by the MICIU/AEI/10.13039/501100011033. P.E. acknowledges support from FPI\_045\_2022, Conselleria d'Educaci\'o i Universitats, Govern de les Illes Balears.

\printcredits

\bibliographystyle{cas-model2-names}

\bibliography{SynthDelays}

\end{document}